\documentclass{article}
\usepackage{spconf,amsmath,graphicx,color,amssymb}
\usepackage{amssymb,relsize}
\usepackage{algorithm,amsmath}
\usepackage[noend]{algpseudocode}
\usepackage{graphicx}
\usepackage{subcaption}

\usepackage{xcolor,colortbl}

\title{sketch based Reduced Memory Hough Transform}
\name{Levi Offen \& Michael Werman\thanks{This research was supported by the Israel Science Foundation and  the  Israel Ministry of Science and Technology.}}
\address{Computer Science\\
	The Hebrew University of Jerusalem\\
	Jerusalem, Israel}%
\begin{document}
\maketitle

\begin{abstract}

This paper proposes using sketch algorithms to represent the votes in Hough transforms.
Replacing the accumulator array with a sketch (Sketch Hough Transform - SHT) significantly reduces the memory needed to compute a Hough transform.
We also present a new sketch, Count Median Update,
which works better than known sketch methods for replacing the accumulator array in the Hough Transform.

\end{abstract}

\section{Introduction}
\label{sec:intro}

Extracting shapes from images
is a key issue in vision and image processing.
 Object detection, especially line detection, is a fundamental operation used in a wide range of applications.
 
The Hough transform \cite{C0}, HT, and the Generalized Hough Transform \cite{ballard_generalizing_1981}, GHT, are  tools based on a voting scheme where image elements vote for parameters of the geometric object. Unfortunately, 
these methods have large memory and computation time requirements
as the parameter space increases exponentially with the dimension of the problem, the number of parameters. On the other hand, reducing the image or the parameter space by quantization  significantly lowers accuracy.

Sketches as methods to approximate frequencies have been successfully used in big data and streaming, where massive data needs to be processed in memory and time efficient manner \cite{charikar_finding_2004,cormode_improved_2005,alon_space_1999}. Sketch algorithms refer to a class of streaming algorithms that represent a large dataset with a compact summary, typically much smaller than the full size of the input. 

One of the problems solved using sketches is the 'frequent items' problem. Given an  sequence of items, find all items whose frequency ('vote value') exceeds a specified fraction of the total number of items: 
A wide variety of algorithms and heuristics have been proposed for this problem, based on sampling, hashing, and counting (see \cite{cormode_finding_2009,liu_methods_2011} for surveys).

\subsection{Our Contribution}\label{sec:contri}
\begin{figure}[ht!] 
\newcommand{\file}{48}
\newcommand{\noise}{05}
\newcommand{\width}{1.2in}
\newcommand{\vsp}{0.2cm}
\newcommand{\hig}{1.4in}
\newcommand{\myspace}{-3mm}
  \hspace*{-0.40cm} 
  \setlength{\tabcolsep}{1pt}
  	\begin{tabular}{ccc}
      		Image & Sketch HT & Classic HT\\
     	\includegraphics[ width=\width, height=\hig]{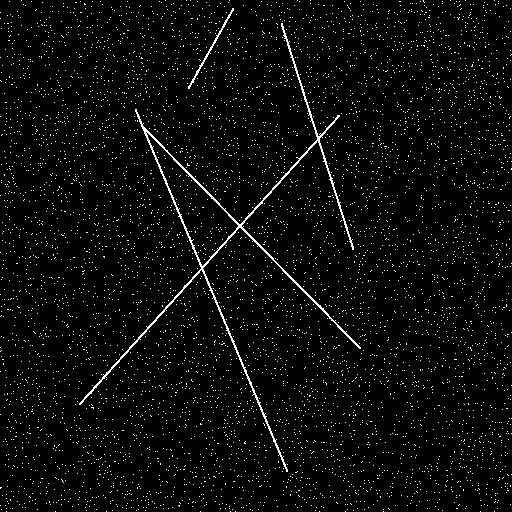} &
     	\includegraphics[ width=\width, height=\hig]
         {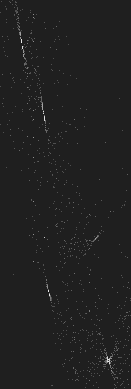} 
        & 
        \includegraphics[width=\width, height=\hig]
        {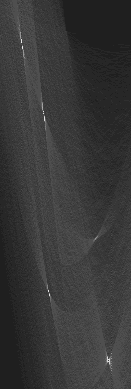}
        \\

  	
  	\end{tabular}
	\caption{The motivation for SHT. Comparing the votes (in parameter space $O$) of the Classic HT to the 'top-frequent-votes' on Sketches HT. It can be seen that many small votes on the right image are omitted on the middle image, while votes around the peaks still preserved. 
    } 
	\label{fig:examples_fake_acc}
\end{figure}
\begin{figure*}[ht] 
\newcommand{\file}{11}
\newcommand{\hash}{9}
\newcommand{\length}{}
\newcommand{\wid}{1.2in}
\newcommand{\filet}{2}
\newcommand{\hig}{1.2in}
\newcommand{\myspace}{-3mm}
\hspace*{-0.62cm} 
\centering
  	\begin{tabular}{cccccc}
  		Classic HT & CU & CM-CU & COUNT& COUNT-CU & COUNT-MU \\
     	\includegraphics[trim=0 0 0 0, clip, width=\wid, height=\hig]{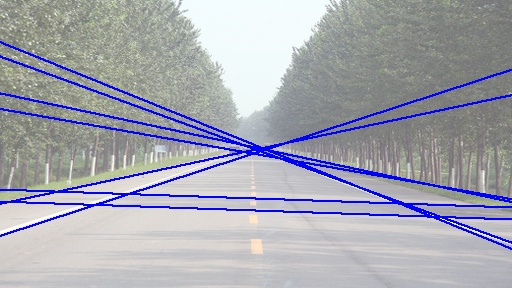} &
     	   \hspace{\myspace}
     	\includegraphics[trim=0 0 0 0, clip, width=\wid, height=\hig]{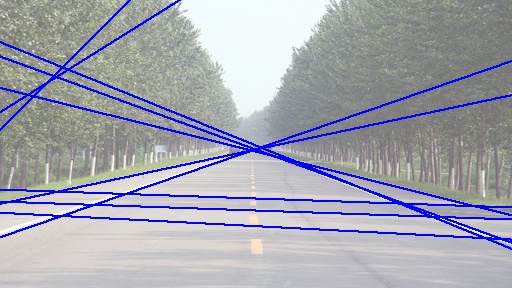} &
     	   \hspace{\myspace}
     	\includegraphics[trim=0 0 0 0, clip, width=\wid, height=\hig]
        {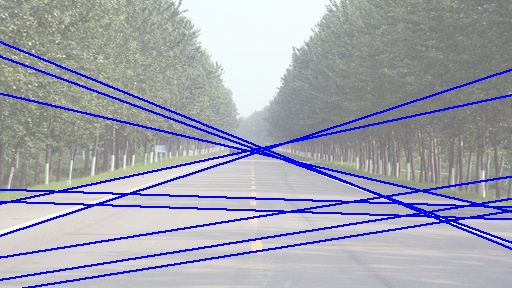} &
     	   \hspace{\myspace}
        \includegraphics[trim=0 0 0 0, clip, width=\wid, height=\hig]{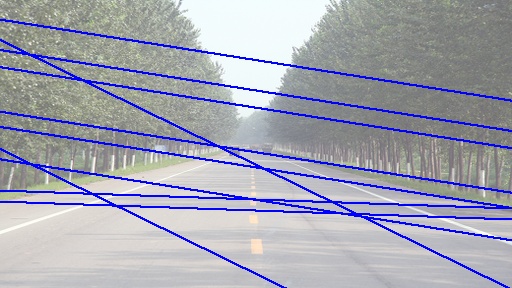}  &
     	   \hspace{\myspace}
        \includegraphics[trim=0 0 0 0, clip, width=\wid, height=\hig]{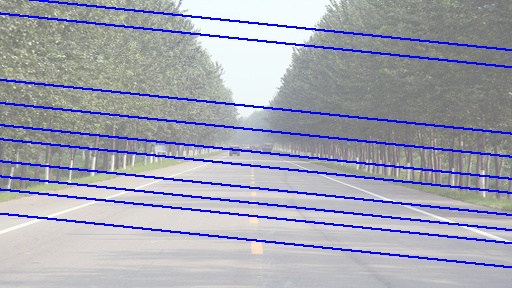}  &
     	   \hspace{\myspace}
        \includegraphics[trim=0 0 0 0, clip, width=\wid, height=\hig]
        {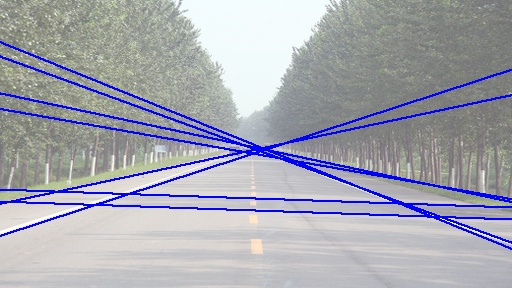}\\  
   	        
        \includegraphics[trim=0 0 0 0, clip, width=\wid, height=\hig]{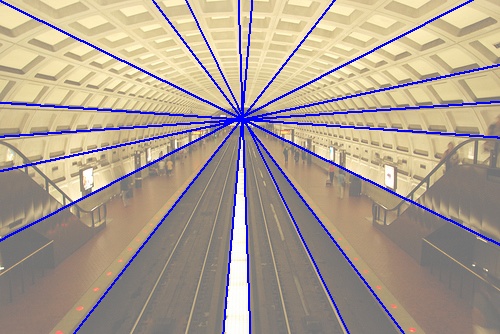} &
     	   \hspace{\myspace}
     	\includegraphics[trim=0 0 0 0, clip, width=\wid, height=\hig]{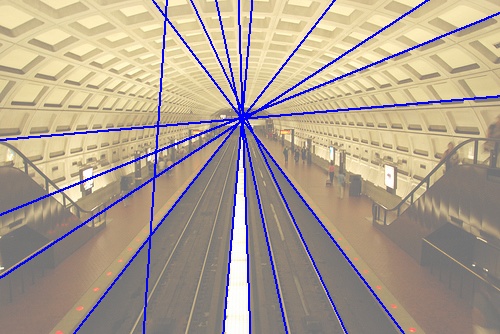} &
     	   \hspace{\myspace}
     	\includegraphics[trim=0 0 0 0, clip, width=\wid, height=\hig]
        {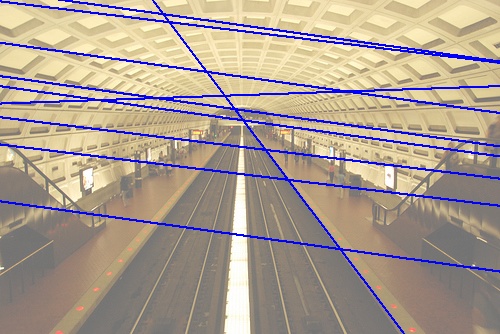} &
     	   \hspace{\myspace}
        \includegraphics[trim=0 0 0 0, clip, width=\wid, height=\hig]{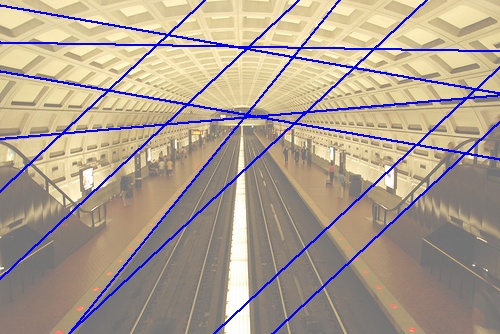}  &
     	   \hspace{\myspace}
        \includegraphics[trim=0 0 0 0, clip, width=\wid, height=\hig]{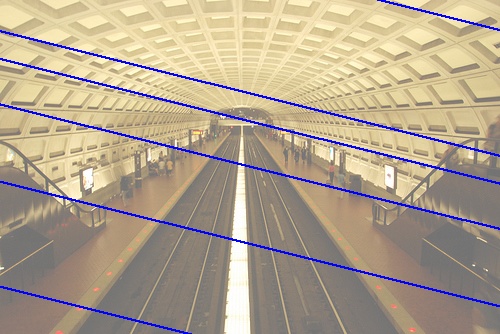}  &
     	   \hspace{\myspace}
        \includegraphics[trim=0 0 0 0, clip, width=\wid, height=\hig]
        {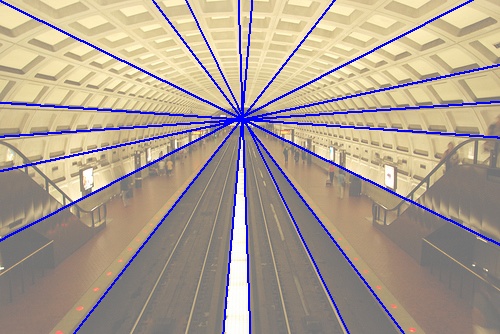}
  	\end{tabular}
	\caption{Detected lines in an image. Classic Hough Transform and various Sketch Hough Transforms.
    } 
	\label{fig:examples_real}
\end{figure*}

Hough transform algorithms detect objects by searching for a local peak in the object parameter space $O$, by the following major steps: 
1. Image  elements, $e$, vote for the cells  in $O$ that agree with $e$. 
2. Local maxima, peaks, in the accumulator, are the detected shapes where the votes are stored in an
 array with the dimension of $O$. We propose replacing the accumulator array with a much smaller sketch.


The 'frequent items' (or 'heavy hitters') problem is not exactly what is needed for the 'peak detection' in Hough transforms, for several reasons: 
\begin{itemize}
\item Due to geometric quantization and noise there are several points related to an object and we only want one.
\item  We want to recognize the objects in the image and are not usually interested in their exact number of votes.
\item There is often a significant amount of noise in the image, which should be ignored.
\end{itemize}

In this paper we show how a sketch algorithms for the 'frequent items' problem can improve Hough transform algorithms, using much less memory and with a better robustness to noise. The main idea is that the votes are only approximated and the peak detection is carried out only around the 'frequent votes'. Figure \ref{fig:examples_fake_acc} shows the difference between the full object parameter space accumulator in Classic HT and the 'top-frequent-votes' of the Sketch HT.   

We also propose a new sketch algorithm,  \textbf{Count Median Update}, which improves the estimation accuracy compared to known methods.

\subsection{Previous Work}
\label{ssec:subcontri}
\subsubsection{Hough Transform Algorithms}
\label{sssec:subsubcontrihough}
The main disadvantages of HT are long computation time and large data storage. Many implementations have being proposed to alleviate these issues \cite{mukhopadhyay_survey_2015}. 

There are probabilistic methods to speed up the HT such as Probabilistic HT \cite{kiryati_probabilistic_1991}, PHT, and Randomized HT \cite{xu_randomized_1993,guo_adaptive_2006,bergen_probabilistic_1991}, RHT.
Although RHT and PHT are computationally  
fast, they are sensitive to noise and occlusions, since the noise pixels have extra impact on these randomized algorithms \cite{lu_detection_2008,guo_probabilistic_2008}. These algorithms use randomness for choosing points in the image space, while we suggest using randomness in parameter space. Our algorithms can also be combined with the previous random methods.
\subsubsection{Data Streaming Sketch Algorithms}

\label{sssec:subsubcontrisketch}
Sketches are concise data summaries of a high-dimensional vector which can be used to estimate queries on it. 
The sketch is a linear projection of the input vector with random vectors defined by hash functions.
Increasing the range of the hash functions co-domain ($w$) increases the accuracy of the estimation, and increasing the number of hash functions decreases the probability of a bad estimate. 
\newcommand{\incr}{INCREMENT}
\newcommand{\query}{QUERY}

The sketch is a $d \times w$ array $C$, and supports $\incr$
and $\query$, 
which can be used for solving the 'frequent items' problem. We outline two sketching approaches
\begin{itemize}
\begin{item} 
\textbf{Count sketch}\cite{charikar_finding_2004} (COUNT) (aka AMS\cite{alon_space_1999}) - There are 2 hash functions per row; $h_i(x)$ and  $s_i(x)$ mapping input items onto $[w]$ and $\{+1,-1\}$ respectively. 
\begin{flalign*}
&\incr(x):\\   
&\quad 
\displaystyle\mathop{\mathlarger{\mathlarger{\mathlarger{\forall}}}}_{1\leq i \leq d} 
\quad C[i,h_i(x)] \leftarrow C[i,h_i(x)]+s_i(x)\\  
&\query\left(x \right): \\  
&\quad \underset{1\leq i \leq d}{\mathrm{median}}\quad C[i,h_i(x)]*s_i(x)
\end{flalign*}
\end{item}
\begin{item}
\textbf{Count Min sketch}\cite{cormode_improved_2005} (CM) - the sketch is similar to $COUNT$ but without the sign hash function. In contrast to $COUNT$ this algorithm returns a biased estimator, overestimating the count. 
\begin{flalign*}
&\incr(x) :\\
&\quad \displaystyle\mathop{\mathlarger{\mathlarger{\mathlarger{\forall}}}}_{1\leq i \leq d}\quad C[i,h_i(x)] \leftarrow C[i,h_i(x)]+1\\
&\query\left(x \right):\\
&\quad\underset{1\leq i \leq d}{\mathrm{min}} \quad  C[i,h_i(x)] 
\end{flalign*} 
\end{item}
\end{itemize}
Let $P$ be the number of items inserted in the sketch, 
$f_i$ the number of times element $i$ was inserted in the sketch ($
\sum_{i}{f_i}=P$), and using $d$ hash functions ($d\times w$ memory),  sketches guarantee:
\\\begin{tabular}{l c c} \label{tab:accuracy_prob}
\emph{Sketch type} & \emph{Estimation Accuracy} & \emph{Success Probability}\\
$COUNT(x)$ & $O\left(\frac{1}{\sqrt[]{w}}\ \sqrt[]{
\sum_{i}\left(f_i\right)^2}\right)$ & $1 - \frac{4}{e^d}$ \\ 
$CM(x)$& $O\left(\frac{1}{w}P\right)$ & $1 - \frac{1}{e^d}$\\
\end{tabular}
Estimation accuracy is a bound on the distance of sketch $\query(x)$  from the real vote value of $x$, $f_x$, and success probability is the probability that this bound fulfilled.

Methods used to improve sketches include:
\begin{itemize}
\item \label{cu} \textbf{Conservative Update} (CU) - conservative updates \cite{estan_new_2002,cohen_spectral_2003} were extended to sketches \cite{cormode_finding_2009,goyal_sketch_2012} to avoid unnecessary updates and reduce overestimation. $C[i,h_i(x)]$ is incremented  only if\footnote{$C[i,h_i(x)]*s_i(x)$ for COUNT} 
$C[i,h_i(x)] \leq \query(x)$.
CU  depends on the order of the incremented items and although  it does not guarantee improvement it often does.

\item \textbf{Lossy counting} (L) - lossy counting \cite{manku_approximate_2002} was extended to sketches\cite{amit_goyal_approximate_2011,goyal_sketch_2012} by removing small votes. 
In this approach, the input is divided into $k$ parts. 
After processing the $t$'th part, small cells $0 < |C[i,j]| \leq t$ (or $\sqrt[]{t}$), are reduced.
In our experiments on images lossy counting did not improve the results
so we do not mention this method again.
\end{itemize}

\label{sketch:solve-topk}Sketches solve the top-k frequent items problem by maintaining a top-k list which is updated during the\break$\incr(x)$ \cite{charikar_finding_2002} or by comparing the $\query(x)$ results for all the $x$'s in parameter space.
\section{new Count sketch}
\label{sec:NCS}
While CM with CU shows significant improvement over CM in many cases, \cite{amit_goyal_approximate_2011} show that CM-CU can reduce the overestimation error by at least 1.5, it appears  that COUNT-CU
\cite{goyal_sketch_2012}
gives little improvement. As CU reduces only over-estimate errors and COUNT also contains under-estimate errors. 

We propose a new variant of conservative update - \textbf{Count Median Update} (COUNT-MU) - that reduces both over and under estimate errors.
\begin{flalign*}
&\incr(x) :\\
&\quad median \leftarrow \underset{1\leq i \leq d}{\mathrm{median}}\quad C[i,h_i(x)]*s_i(x) \\ 
&\quad 
\displaystyle\mathop{\mathlarger{\mathlarger{\mathlarger{\forall}}}}_{1\leq i \leq d} 
\quad \textbf{If} \quad median == C[i,h_i(x)]*s_i(x):\\
&\qquad\qquad\qquad C[i,h_i(x)] \leftarrow C[i,h_i(x)]+s_i(x)
\end{flalign*} 

The motivation for this method is similar to CM-CU, updating $x$ should only affect the cells equal to $\query(x)$\footnote{\label{count-mu-update}updating in the range  $[median\_low,median\_high]$ for even $d$ or $[median-1,median+1]$ for odd $d$, instead of just $median$ slightly improves   results}. The other counters are 'wrong' since they were notably influenced by other elements, so incrementing them will increase  noise and inaccuracy. 

Our new sketch algorithm is significantly better than COUNT-CM and usually  better than CM-CU (see \ref{sec:exp}) and could possibly be of use
in other streaming data/NLP queries.

\section{Sketch Hough Transform (SHT)}
\label{sec:SKHT}

\newcommand{\maxl}{peaks\_num}
\newcommand{\hashn}{d}
\newcommand{\memn}{mem}
\newcommand{\maxrho}{\rho\_max}
We claim that any algorithm which estimates the 'frequent items' can be used to improve Hough transforms algorithms. 

The Hough Transform's parameters are the polar coordinates of the line,
 $\theta \in [0,\pi]$ and $\rho \in R$, which are the angle of the normal to the line and the distance from the origin to the line.  
Let $skt$ be the used sketch, $\hashn$ the number of hash functions used in $skt$, $\memn$ the  memory  used by $skt$ (the size of hash's co-domain$\times d$), $\maxl$ the maximum number of lines we expect to find in the image, $\maxrho$ the maximum distance of a line in the image from the origin,
and $|P|$ the number of edge points in the image. 

The Classic Hough Transform, CHT, stores the line votes in an accumulator array which ranges over all the object space - $\textnormal{\# of angles} \times 2\maxrho$.
Algorithm \ref{alg:sht}, SHT, replaces this accumulator with a smaller sketch,  using 
$ \memn $
memory and with a $\Theta \left( 1- \frac{1}{e^\hashn}\right)$  probability returns a superset of the CHT result. 

\begin{algorithm}
\caption{Sketch Hough Transform}\label{alg:sht}
\begin{algorithmic}[1]
\Procedure{SHT}{$Im,
\maxl,\memn,\hashn$}
\State $P\gets edgePoints(Im)$
\State $w\gets \lceil\frac{\memn}{\hashn}\rceil$
\State $hash\_domain \gets [-\rho\_max,\rho\_max]$ 
\State $top\_list \gets \varnothing$
\For {$\theta$}
\State $skt \gets initSketch(hash\_domain,w,d)$
\For {$(x,y) \in P$}
\State $\rho \gets  x \cos ( \theta ) + y \sin ( \theta ) $
\State $skt . \incr(\rho)$\label{row:1}
\EndFor
\State $top\_list . add (\{\theta, skt . getTops(2*\maxl)\})$\label{row:2}
\EndFor
\State $peaks \gets $ all $(\theta,\rho)$ which are peaks
in $top\_list$\label{row:3} 
\State \textbf{return} the $2*\maxl$ elements with the most votes in $peaks$\label{row:4}
\EndProcedure
\end{algorithmic}
\end{algorithm}
$initSketch(hash\_domain,w,d)$  returns a sketch which stores votes for elements from $hash\_domain$ using $d$ hash functions to a size $w$ co-domain -  using $w \times  d$ memory.
$getTops(k)$ returns the top-$k$ 'frequent items' from the sketch and is calculated at line \ref{row:1} with $O\left(\maxl\right)$ memory.

$top\_list$ size can be limited to $O \left(\maxl \right)$ by searching locally for peaks within windows of a fixed number of angles.

Although  the number of votes for a line in SHT is different (depending on $|P|$, $\memn$, $\hashn$ and $skt$)
from the  number of votes in CHT, using the right configuration for $mem,\hashn$ (see \ref{ssec:expu_synth}) results in a superset of lines in almost the same order (sorted by votes) as CHT. Additionally,  a simple check can remove false lines which do not exist in the image.

\newcommand{\pltmemchange}{(b) }
\newcommand{\pltnoisechange}{(a) }
\newcommand{\plthashchange}{(c) }

\begin{figure*}[ht!] 
\newcommand{\file}{48}
\newcommand{\noise}{05}
\newcommand{\width}{0.24}
\newcommand{\vsp}{0.1cm}
\newcommand{\wid}{0.35\linewidth}
\newcommand{\hig}{0.25\linewidth}
\newcommand{\myspace}{-4mm}
\newcommand{\nolable}{_no_lable}

\hspace*{-1cm} 
\centering
  	\begin{tabular}{ccc}
     	\includegraphics[trim=0 0 0 0, clip, width=\wid, height=\hig]{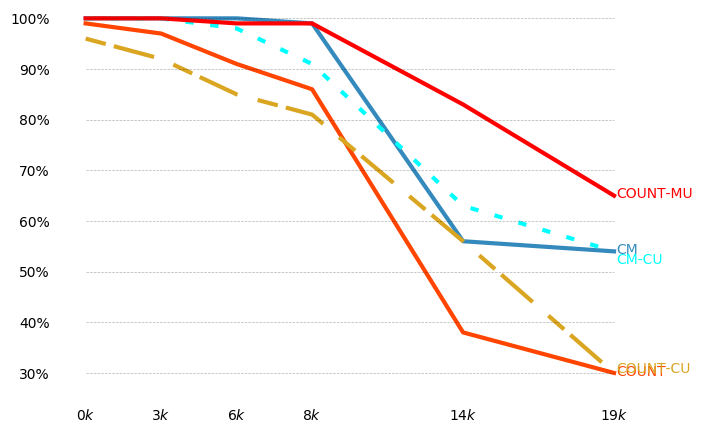}\label{fig:exp_plots:noise} &
     	   \hspace{\myspace}
     	        \includegraphics[trim=0 0 0 0, clip, width=\wid, height=\hig]{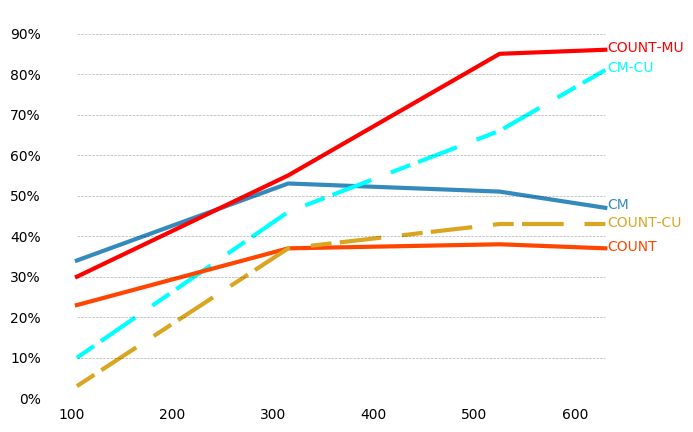}\label{fig:exp_plots:mem}   &
     	   \hspace{\myspace}
			\includegraphics[trim=0 0 0 0, clip, width=\wid, height=\hig]{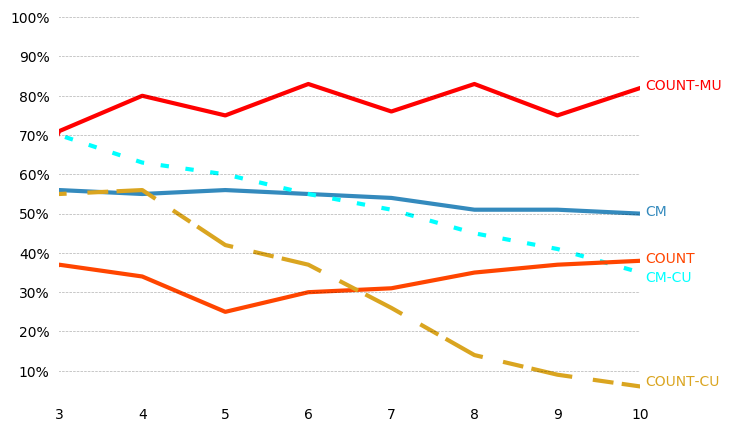}\label{fig:exp_plots:hash}
     	   \hspace{\myspace}\\
        \pltnoisechange &\pltmemchange &\plthashchange
  	\end{tabular}
	\caption{
    \pltnoisechange Quality as a function of  of noise in 1000's of edge elements.
    \\\pltmemchange Quality as a function of memory.
    } 
    \plthashchange Quality as a function of number of hash functions.
	\label{fig:exp_plots}
\end{figure*}

\section{Experiments}
\label{sec:exp}
In order to show the effectiveness of the proposed Sketch Hough Transform (SHT) we run it with several sketching methods: Count Min (CM), Count Min with conservative update (CM-CU),
Count (COUNT), Count with conservative update (COUNT-CU),
and Count Median Update (COUNT-MU), our method.
All of them with and without lossy counting.

The accuracy of SHT results were calculated by comparing them to the result of CHT using an accumulator of $180 \times 1024$ (184320) memory. Each algorithm was run 10 times on an image and the results quality (mean accuracy) is reported. 
\subsection{Synthetic Line Images}
\label{ssec:expu_synth}
We created 204 synthetic images ($512\times 512$) containing 1-5 random lines and added uniform noise. We ran SHT algorithms 10 times on each image, and the quality of the results were compared to the result of the classic Hough Transform. As lossy counting did not have a significant effect on the sketches in these cases, we don't show lossy counting results. 

Plot \ref{fig:exp_plots}\pltnoisechange shows the dependence of SHT quality on the amount of noise for $275$ bytes of memory. It can be clearly seen that the results of COUNT-MU are superior to all the other sketches. The advantage of our method increases with the noise.


Plot \ref{fig:exp_plots}\pltmemchange shows the dependence of SHT quality on sketch memory size (hash co-domain$\times$number of hash functions) for images with 19k noise points. It can be seen that CM has better results than CM-CU for sketches with small memory, while CM-CU is better with memory size above 420. Our method, COUNT-MU, is superior to all other methods.

Plot \ref{fig:exp_plots}\plthashchange shows the dependence of SHT quality on the number of hash functions, using  $210$ bytes of memory. The seesaw pattern  in COUNT-MU is a result of the difference in the definition of median for an even or odd number of elements,  number of hash functions.


\subsection{Real Images}
\label{ssec:exp_real}
We ran the SHT on 15 random real images of various sizes containing roads, train tracks, skylines and landscapes. Figure~\ref{fig:examples_real}  shows the detected lines on an image for Classic HT and SHT with various sketch types.


\begin {table}[H]
\caption {SHT accuracy using $275$ bytes of memory is:}
\begin{center}
\definecolor{Gray}{gray}{0.75}
\begin{tabular}{|c|c|}
\hline
Sketch&Quality\\
\hline  \hline  
CM &76\%\\
CM-CU &90\%\\
COUNT&56\%\\
COUNT-CU&20\%\\
\cellcolor{Gray}{\textbf{COUNT-MU}}&\cellcolor{Gray}{\textbf{96\%}}\\
\end{tabular}
\end{center}
\end{table}

It can be clearly seen  that the results of COUNT-MU, our method, are superior to all other sketches on real images too. 

\section{Conclusion}
\label{sec:concl}
We introduced the Sketch Hough Transform, SHT, algorithm 
that reduced the amount of  memory and increased the robustness to noise compared to the Classic Hough Transform. We showed that the results of SHT, using a small memory are almost the same as the classic Hough Transform. 

We also proposed a new sketch, Count Median Update, and showed that this new sketch is significantly superior to other sketching methods especially as the noise in the image increased.

\pagebreak

\bibliographystyle{IEEEbib}
\bibliography{Zotero,refs}

\end{document}